\documentclass[journal]{IEEEtran}
\usepackage{amsmath,amsfonts}
\usepackage{algorithmic}
\usepackage{algorithm}
\usepackage{array}
\usepackage[caption=false,font=normalsize,labelfont=sf,textfont=sf]{subfig}
\usepackage{textcomp}
\usepackage{stfloats}
\usepackage{url}
\usepackage{verbatim}
\usepackage{graphicx}
\usepackage{cite}
\usepackage{balance}
\usepackage{multirow}
\usepackage{booktabs}
\usepackage{tabularx}
\newcommand{\T}{\mathrm{T}}

\begin{document}

\title{Selectivity Matters: Source Node Influence Pruning \\for Unsupervised Graph Domain Adaptation}

\author{Ridong Han,~\IEEEmembership{Member,~IEEE,} Yawen Shen, Zhongnian Li, Tongfeng Sun, Xinzheng Xu, \\and Abdulmotaleb El Saddik,~\IEEEmembership{Fellow,~IEEE}
        %<-this % stops a space
\thanks{Ridong Han, Yawen Shen, Zhongnian Li, Tongfeng Sun, and Xinzheng Xu are with the School of Computer Science and Technology / School of Artificial Intelligence, China University of Mining and Technology, Xuzhou 221116, China, and with the Mine Digitization Engineering Research Center of the Ministry of Education, Xuzhou 221116, China, and also with the Jiangsu Provincial Industrial Technology Engineering Center for Intelligent Sensing
and Emergency IoT in Underground Space, Xuzhou 221116, China (e-mails: \{hanrd, ts24170117p31, zhongnianli, suntf, xxzheng\}@cumt.edu.cn).}% <-this % stops a space
\thanks{Abdulmotaleb El Saddik is with the School of Electrical Engineering and
Computer Science, University of Ottawa, Ottawa, Ontario KiN5N6, Canada
(email: elsaddik@uottawa.ca).}
\thanks{Corresponding author: Xinzheng Xu (e-mail: xxzheng@cumt.edu.cn).}
}

\markboth{This work is a preprint and has been submitted for peer review.}%
{R. Han, Y. Shen \MakeLowercase{\textit{et al.}}: Selectivity Matters: Source Node Influence Pruning for Unsupervised Graph Domain Adaptation}


\maketitle

\begin{abstract}
Unsupervised Graph Domain Adaptation (UGDA) aims to facilitate knowledge transfer from a labeled source graph to an unlabeled target graph by mitigating cross-domain distribution shifts. 
Existing methods primarily focus on node-level feature alignment in latent spaces, 
relying on the implicit assumption that all source nodes contribute positively to the alignment.
However, this assumption often fails because a node's semantic information is intrinsically coupled with its topological graph structure. 
Due to structural shifts, source nodes with severe structural deviations (e.g., structural outliers) lack semantic counterparts in the target graph, 
and forcing alignment on them introduces severe noise and causes negative transfer.
To bridge this gap, we argue that selective source node utilization is superior to full-graph training, thereby shifting the research paradigm from feature-level alignment to data-level refinement. To this end, we propose Source Node Influence Pruning (SNIP), a novel model-agnostic, data-centric refinement framework. 
Specifically, SNIP quantifies the structural discrepancy between individual source nodes and the target domain by integrating multiple centrality measures, assigning each source node an influence score. 
A rank-based normalization mechanism is further employed to eliminate scale variations across different measures, allowing SNIP to effectively identify and filter out structurally incompatible nodes with low influence scores. 
As a plug-and-play method, SNIP constructs a refined ``sub-source'' graph that is inherently more beneficial for subsequent alignment.
Comprehensive experiments across eight transfer scenarios on five real-world datasets demonstrate that SNIP consistently outperforms competitive baselines and significantly enhances adaptation performance, validating the superiority of selective node utilization over full-graph training. 
The code is available at \url{https://github.com/RidongHan/SNIP}.
\end{abstract}


\section{Introduction}
\IEEEPARstart{T}{he} widespread adoption of Graph Neural Networks (GNNs) \cite{Kipf2017Semi,Yang2023Graph} in social network analysis, bioinformatics, and recommender systems has highlighted the necessity of addressing distributional discrepancies in graph-structured data.
These discrepancies, arising from varying data acquisition protocols or evolving topological structures, severely degrade model generalization.
Unsupervised Graph Domain Adaptation (UGDA) \cite{Shi2025Domain,Wilson2020A} aims to mitigate these distributional shifts, facilitating knowledge transfer from a labeled source graph to an unlabeled target graph.

\begin{figure}[!tbp]
	\centering
	\includegraphics[width=0.80\linewidth]{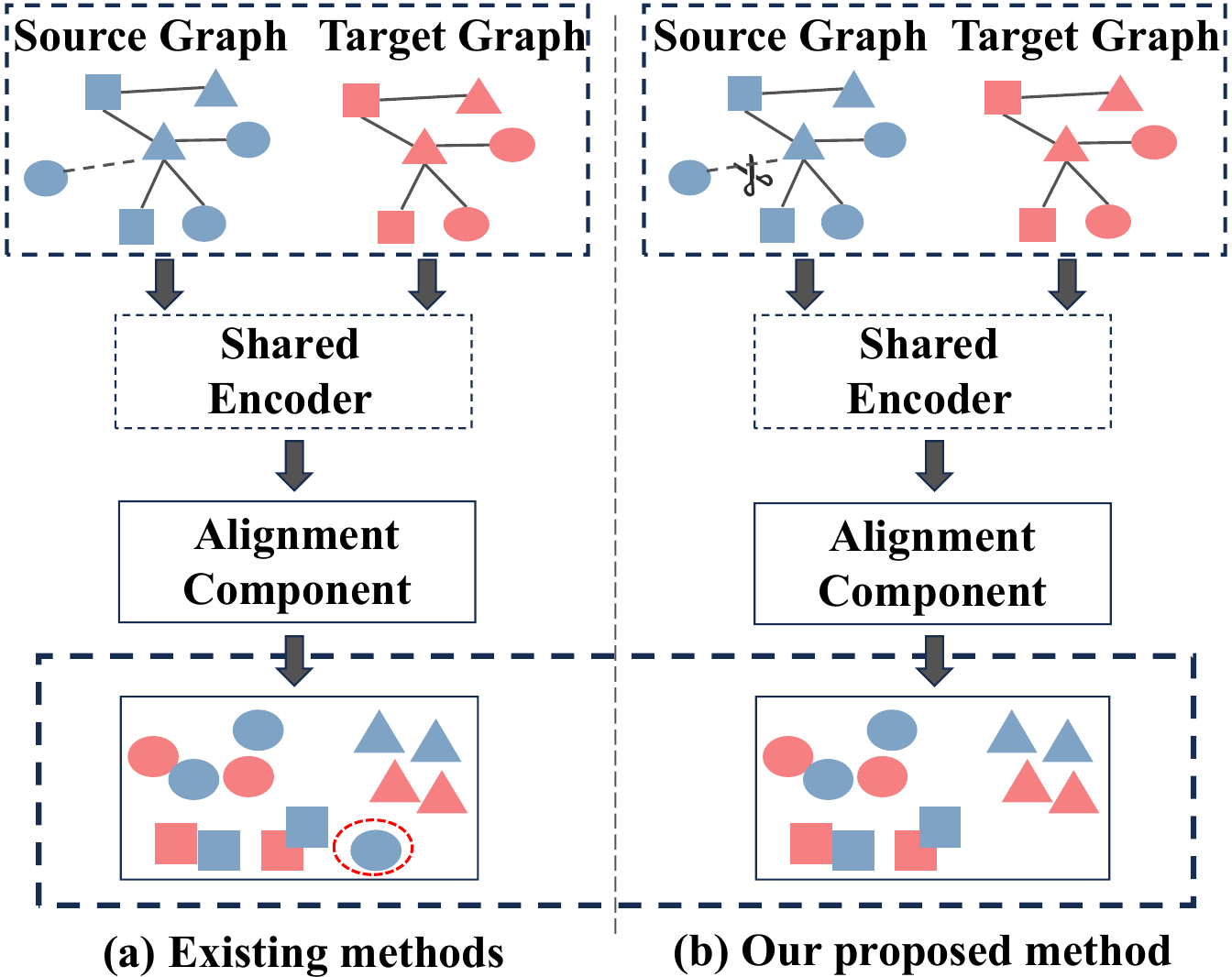}
	\caption{Conceptual comparison between existing feature-level alignment methods and our proposed pruning approach. The red dashed ellipse highlights how structural outliers distort the shared feature space and cause negative transfer, which is mitigated by our selective source node utilization.}
    \label{fig:figure-illustration-difference}
\end{figure}

Existing UGDA methods primarily focus on \textit{feature-level alignment} in latent spaces, typically employing regularization techniques such as Maximum Mean Discrepancy (MMD) \cite{Shen2021Network}, graph subtree discrepancy \cite{Wu2023Non}, or adversarial learning \cite{Xiao2023SPA} to enforce cross-domain consistency.
While effective at mitigating feature-level shifts, these methods fundamentally rely on the implicit assumption that \textit{all source nodes contribute positively to cross-domain knowledge transfer}, which is often invalid in practice.
By treating the source graph as a whole, these methods attempt to force a global alignment with the target domain. However, in graph-structured data, a node's semantic information is intrinsically coupled with its topological structure. 
Due to pronounced structural shifts, 
source nodes exhibiting extreme topological properties, such as highly central hubs or structural outliers, often lack corresponding semantic counterparts in the target domain.
Consequently, forcing the alignment on these structurally incompatible nodes inevitably injects significant noise, thereby distorting cross-domain representations and exacerbating negative transfer.

As illustrated in Fig.~\ref{fig:figure-illustration-difference}, existing feature-level alignment methods typically utilize the entire source graph for alignment, which are highly vulnerable to the structurally incompatible nodes.
For instance, outlier nodes in the feature space (circled in red dashed ellipse) can easily disturb the alignment process and cause the shared encoder to learn distorted representations for the target domain, degrading adaptation performance. 
In contrast, we argue that a ``clean'' source domain is a fundamental prerequisite for effective adaptation. In other words, \textbf{selective source node utilization is superior to full-graph training}. 
Guided by this philosophy, our goal is to identify and prune source nodes that are topologically incompatible with the target structure, thereby establishing a robust foundation for subsequent feature-based alignment.

To this end, we shift the research paradigm from conventional \textit{feature-level alignment} to \textit{data-level refinement}, moving from the question of ``\textit{how to align}'' to ``\textit{what to align}''. 
In this paper, 
we propose a novel data-centric refinement strategy named \textbf{S}ource \textbf{N}ode \textbf{I}nfluence \textbf{P}runing (SNIP).
Instead of indiscriminately utilizing all source nodes, SNIP actively filters out structural noise by leveraging the target domain's global topological statistics as a reference benchmark. 
To achieve this, SNIP first quantifies the structural discrepancy between individual source nodes and the target domain to compute node influence scores, by integrating multi-dimensional centrality measures such as degree centrality and betweenness centrality.
To mitigate the intrinsic scale variations among multiple centrality measures, a rank-based normalization mechanism is introduced, which converts the absolute values of each centrality measure into their corresponding relative percentile ranks within the source graph, thereby mapping diverse centrality measures into a uniform, non-parametric ordinal scale.
Subsequently, SNIP ranks all source nodes based on their calibrated influence scores (i.e., the scores processed via the rank-based normalization mechanism), and filters out structurally incompatible source nodes with low influence scores.
This process effectively eliminates source nodes that diverge significantly from the target domain's profile, 
constructing a refined ``sub-source'' graph that is inherently more beneficial to cross-domain alignment.

Notably, since SNIP operates only at the data level and remains entirely model-agnostic, it can serve as a seamless, plug-and-play module for any existing UGDA method.
We conduct extensive experiments using multiple backbone models on five real-world datasets, comprising three citation networks (i.e., ACMv9, Citationv1, and DBLPv7) and two Twitch gamer networks (i.e., Germany and England). 
Results demonstrate that SNIP consistently outperforms competitive baselines, significantly enhancing adaptation performance. 
In particular, across six transfer scenarios on three citation networks, SNIP improves AdaGCN by an average of 3.01\% in Micro F1 and 4.62\% in Macro F1, while achieving gains of 1.32\% and 2.23\% with A2GNN, respectively.
Subsequently, ablation studies confirm the necessity and effectiveness of each component within SNIP. Furthermore, sensitivity analyses validate that SNIP maintains high robustness across a wide range of hyper-parameters, while embedding visualizations offer direct qualitative evidence that our data-centric refinement effectively mitigates representation distortion and eliminates negative transfer.
Our contributions are summarized as follows:

\begin{itemize}
	\item \textbf{Data-Centric Refinement}: We challenge the conventional ``align-all'' paradigm, and propose an active source domain refinement framework. By identifying and pruning structural outliers, as a primary source of negative knowledge transfer, we shift the focus from ``how to align'' to ``what to align''.
    \item \textbf{Plug-and-Play Method}: We introduce a lightweight pruning strategy named SNIP, which can be seamlessly integrated into existing baselines to enhance their performance with minimal computational overhead.
	\item \textbf{Empirical Validation}: Extensive experiments on five public datasets demonstrate that pruning a calculated proportion of structurally incompatible source nodes yields significant improvements, thereby confirming the necessity of structural filtering in graph domain adaptation.
\end{itemize}

\section{Related Work}
\noindent
\textbf{Unsupervised Graph Domain Adaptation}. 
With the wide application of graph neural networks, Unsupervised Graph Domain Adaptation (UGDA) has become a key technology for addressing cross-domain distribution shifts in graph-structured data.
The field has evolved from basic feature alignment to more sophisticated structural integration and theory-driven frameworks.
Early studies primarily focused on learning domain-invariant node representations.
UDAGCN~\cite{Wu2020Unsupervised} introduced adversarial training to align node-level representations across domains, 
while ACDNE~\cite{Shen2020Adversarial} used dual-stream feature extractors to jointly capture node attributes and local topological properties.
ASN~\cite{Zhang2021Adversarial} further employed adversarial separation networks to extract domain-invariant features.

Subsequent studies increasingly emphasized the role of graph structure in domain adaptation.
JHGDA \cite{Shi2023Improving} explored the impact of graph hierarchy on class-conditional distributions, 
and GRADE \cite{Wu2023Non} derived generalization error bounds based on the Weisfeiler-Lehman operator and used them to guide subtree-level alignment. 
More recent studies have further explored how graph propagation, structural properties, and target-domain representations affect cross-domain generalization.
A2GNN~\cite{Liu2024Rethinking} demonstrated the effectiveness of asymmetric message propagation in the source and target domains. 
PairAlign~\cite{Liu2024Pairwise} decoupled the alignment of structural and label shifts. 
TDSS~\cite{Chen2025Smoothness} improved target-side smoothness through Laplacian smoothing and neighborhood sampling, thereby reducing structural discrepancies. 
TASSC~\cite{zou2025target} integrated topological and semantic consistency to enhance target representation learning, 
while DAGNN~\cite{yang2026dagnn} extended asymmetric graph adaptation~\cite{Liu2024Rethinking} by disentangling domain-invariant and domain-specific representations and introducing topology-aware distribution alignment.
In addition to these model-centric approaches, 
GraphAlign~\cite{huang2024can-modifying} explored data-centric UGDA by generating a completely new, compact, and transferable graph.
Following the principles of distribution alignment and graph rescaling, It jointly optimizes the features and topology of the generated graph to reduce its discrepancy from the target graph, while preserving source-domain knowledge through gradient matching and graph regularization.

Despite these advances, existing methods generally treat source nodes as equally transferable and therefore overlook the heterogeneous contributions of individual source nodes to cross-domain transfer.
To fill this gap, we propose SNIP, a source node influence pruning framework that evaluates node-level structural compatibility and removes source nodes that are structurally incompatible with the target domain.
Unlike GraphAlign~\cite{huang2024can-modifying}, which synthesizes a new graph by jointly optimizing its features and structure, SNIP preserves authentic source-domain data and selectively retains a subset of source nodes together with their topology. 
By explicitly quantifying the structural compatibility of individual source nodes, SNIP provides a lightweight, interpretable, model-agnostic, and plug-and-play data refinement strategy rather than a graph generation or condensation method.

\vspace{1ex}
\noindent\textbf{Node Centrality}. 
Node centrality \cite{Park2026Centrality} is a key metric in network science used to quantify the importance of nodes in complex networks.
Common indicators, such as degree \cite{Brodka2011A}, betweenness \cite{Maurya2019Fast}, closeness \cite{Liu2023Closeness}, and eigenvector \cite{Frost2023Eigenvector} centrality, describe a node's structural influence across multiple dimensions, ranging from local connectivity to global path dependence.
In Graph Neural Networks (GNNs), centrality metrics is commonly used for positional encoding and enhancing attention mechanisms \cite{Li2024Centrality}.
However, this study adopts a distinct approach: rather than merely utilizing centrality to identify key nodes for representation enhancement, we treat it as a structural distribution detector.
By measuring the structural discrepancy between source nodes from the average structural characteristics of the target domain, we quantify the transfer difficulty of nodes in the structural dimension.
This provides a theoretical basis for subsequent pruning strategies.

\vspace{1ex}
\noindent\textbf{Graph Pruning}. Graph pruning \cite{lin2026knowledge} has traditionally been used to reduce computational complexity and eliminate random noise. 
Advances in graph augmentation techniques have led to methods like DropEdge and graph sparsification \cite{Rong2020DropEdge}, which effectively mitigate over-smoothing and enhance model generalization.
However, existing pruning schemes primarily rely on random dropping or simple attention-based weights.
These methods are developed without considering domain shift.
This oversight may result in the pruning of nodes essential for cross-domain alignment and the retention of nodes that impede the domain adaptation.
By introducing a rank-based normalization mechanism to integrate multi-dimensional structural metrics, we selectively remove source nodes that exacerbate the domain gap. 
Consequently, this approach promotes more robust structural-level alignment across domains.

\section{Preliminaries}
\noindent
\textbf{Notations.} 
Let $\mathcal{G}=(\mathcal{V},\mathcal{E})$ denotes an undirected, unweighted graph comprising $n=|\mathcal{V}|$ nodes and $m=|\mathcal{E}|$ edges, where $\mathcal{V}$ and $\mathcal{E}$ represent the node set and edge set, respectively. 
We present the graph data as a tuple $\mathcal{G}=(\mathbf{A},\mathbf{X},\mathbf{Y})$. 
Specifically, 
$\mathbf{A} \in \mathbb{R}^{n \times n}$ denotes the symmetric adjacency matrix, where $\mathbf{A}_{i, j}=1$ if an edge $e_{i, j} \in \mathcal{E}$ exists between node $v_{i}$ and $v_{j}$, and $ \mathbf{A}_{i, j}=0$ otherwise. 
Furthermore,
the node attribute matrix is denoted by $\mathbf{X}=[x_1,x_2,\cdots,x_n]^{\T} \in \mathbb{R}^{n \times d}$, where $x_v\in\mathbb{R}^d$ represents the feature vector of node $v\in\mathcal{V}$, and $d$ represents the feature dimension.
Analogously, $\mathbf{Y}=[y_1,y_2,\cdots,y_n]^{\T} \in \mathbb{R}^{n \times C}$ denotes the ground-truth label matrix, with $C$ representing the total number of categories.

\vspace{1ex}
\noindent\textbf{Problem Definition.} 
Unsupervised Graph Domain Adaptation (UGDA) facilitates knowledge transfer from a labeled source graph $\mathcal{G}^s=(\mathbf{A}^s,\mathbf{X}^s,\mathbf{Y}^s)$ to an unlabeled target graph $\mathcal{G}^t=(\mathbf{A}^t,\mathbf{X}^t)$. 
This paradigm typically operates under the generalized covariate shift assumption \cite{David2006analysis,David2009a-thory}: both graphs follow different feature distributions yet share the same conditional distribution of labels, i.e., $\mathbb{P}_s(\mathbf{X}^s)\neq \mathbb{P}_t(\mathbf{X}^t)$ and $\mathbb{P}_s(\mathbf{Y}^s|\mathbf{X}^s) = \mathbb{P}_t(\mathbf{Y}^t|\mathbf{X}^t)$.
Given that ground-truth labels are accessible solely for the source graph $\mathcal{G}^s$, the ultimate objective of UGDA is to collaboratively leverage both $\mathcal{G}^s$ and $\mathcal{G}^t$ to train a robust graph neural network. By effectively mitigating cross-domain distribution shifts, this network is expected to yield high-fidelity node representations, thereby accurately predicting labels $\mathbf{Y}^t$ for the target graph $\mathcal{G}^t$.

\section{Methodology}
This section begins with empirical and statistical analyses substantiating our motivation. 
Guided by the insights obtained, we introduce a data-centric refinement framework termed Source Node Influence Pruning (SNIP). Finally, we provide a theoretical analysis of the proposed approach.

\subsection{Empirical Analysis}
\label{sec:empirical-analysis}

\begin{figure}[!tbp]
	\centering
    \includegraphics[width=0.75\linewidth]{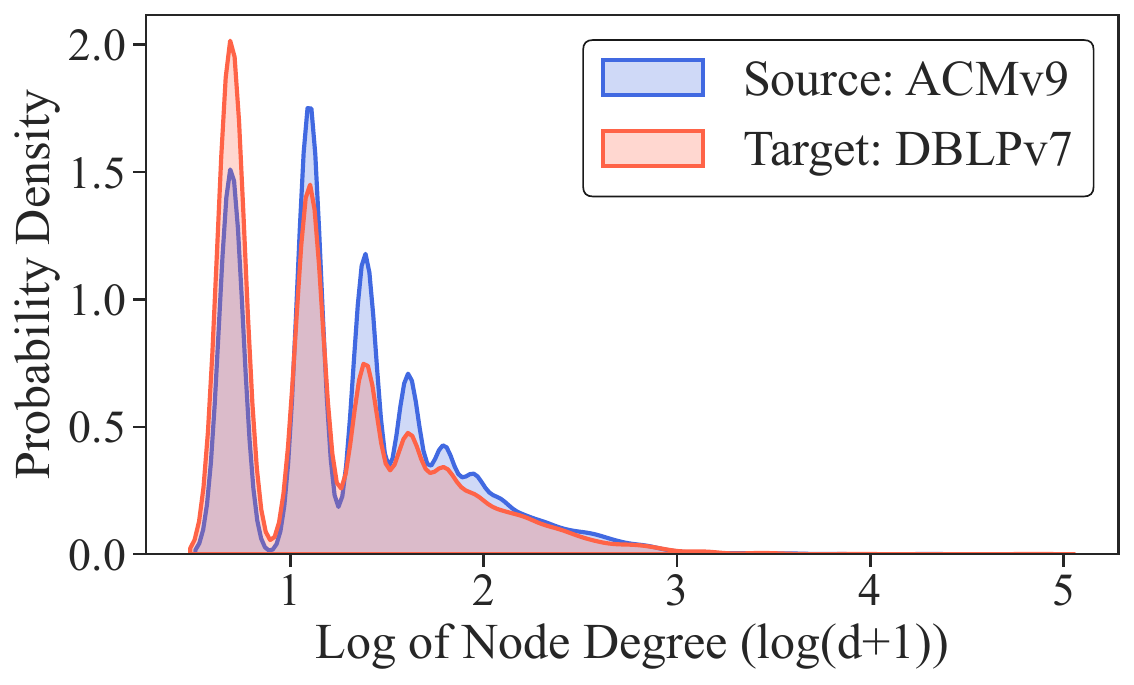}
    \caption{Kernel Density Estimation (KDE) curves of log-transformed node degree distributions across source (ACMv9) and target (DBLPv7) domains.}
    \label{fig:empirical-observation-1}
\end{figure}

\noindent
\textbf{Observation 1: Significant Cross-Domain Distributional Shifts.} 
We investigate cross-domain differences by visualizing the topological properties of the source (ACMv9) and target (DBLPv7) domains.
Fig.~\ref{fig:empirical-observation-1} illustrates the Kernel Density Estimation (KDE) curves of log-transformed node degrees.
It is observable that \textit{the curves of ACMv9 and DBLPv7 exhibit significant deviations in peak positions and overall distribution shapes}, reflecting significant distributional shifts between domains.
Even assuming perfect feature alignment, such structural mismatches persist. 
Enforcing alignment on nodes with deviant structures, such as outliers with extreme degrees, can induce negative transfer.
Consequently, selective source node utilization is essential to retain only source nodes structurally compatible with the target domain.

\begin{figure}[!tbp]
	\centering   
    \includegraphics[width=\linewidth, trim=0 30 0 80, clip]{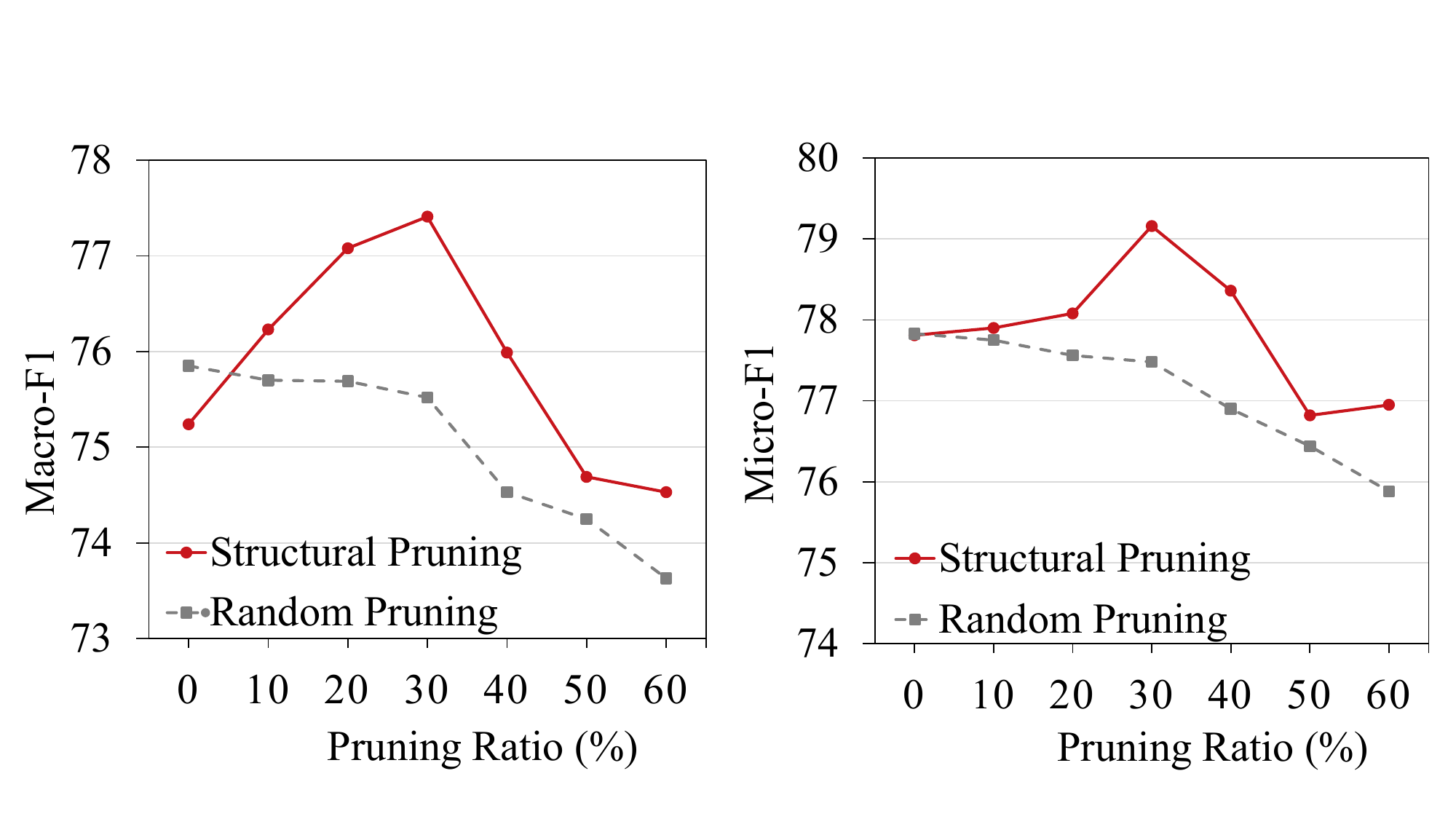}
    \caption{Performance comparison curves between ``Structural Pruning'' and ``Random Pruning'' under different pruning ratios across the source (ACMv9) and target (DBLPv7) domains.}
    \label{fig:empirical-observation-2}
\end{figure}

\vspace{1ex}
\noindent
\textbf{Observation 2: Negative Transfer Caused by Structurally Incompatible Nodes.}
To validate the hypothesis that selective source node utilization is critical, we conduct a preliminary experiment utilizing Degree Centrality as the structural measure to prune a specified proportion of source nodes that exhibit significant discrepancies with the target domain, a strategy referred to as ``Structural Pruning''. Specific model details and experimental settings are provided in Section~\ref{sec:experimental-settings}. 
We compare ``Structural Pruning'' against ``Random Pruning'' under various pruning ratios across the source (ACMv9) and target (DBLPv7) domains, as illustrated in Fig.~\ref{fig:empirical-observation-2}.
Random pruning exhibits performance decay as the pruning ratio increases, reflecting the indiscriminate removal of essential structural information.
Conversely, \textit{structural pruning demonstrates an initial performance improvement followed by a decline and yields improved Micro-F1 and Macro-F1 scores, reflecting the negative transfer phenomenon caused by structurally incompatible nodes}.
Although pruning inevitably sacrifices some structural information, the above results indicate that the benefits of filtering structural noise outweigh the loss of raw topological structure.
These findings empirically support the premise that selective source node utilization is superior to full-graph training. Identifying and filtering out structurally incompatible source nodes significantly improves graph adaptation performance in the target domain.

\subsection{Source Node Influence Pruning Framework}
\label{sec:source-node-influence-pruning}

\begin{figure}[!tbp]
	\centering   
    \includegraphics[width=0.98\linewidth]{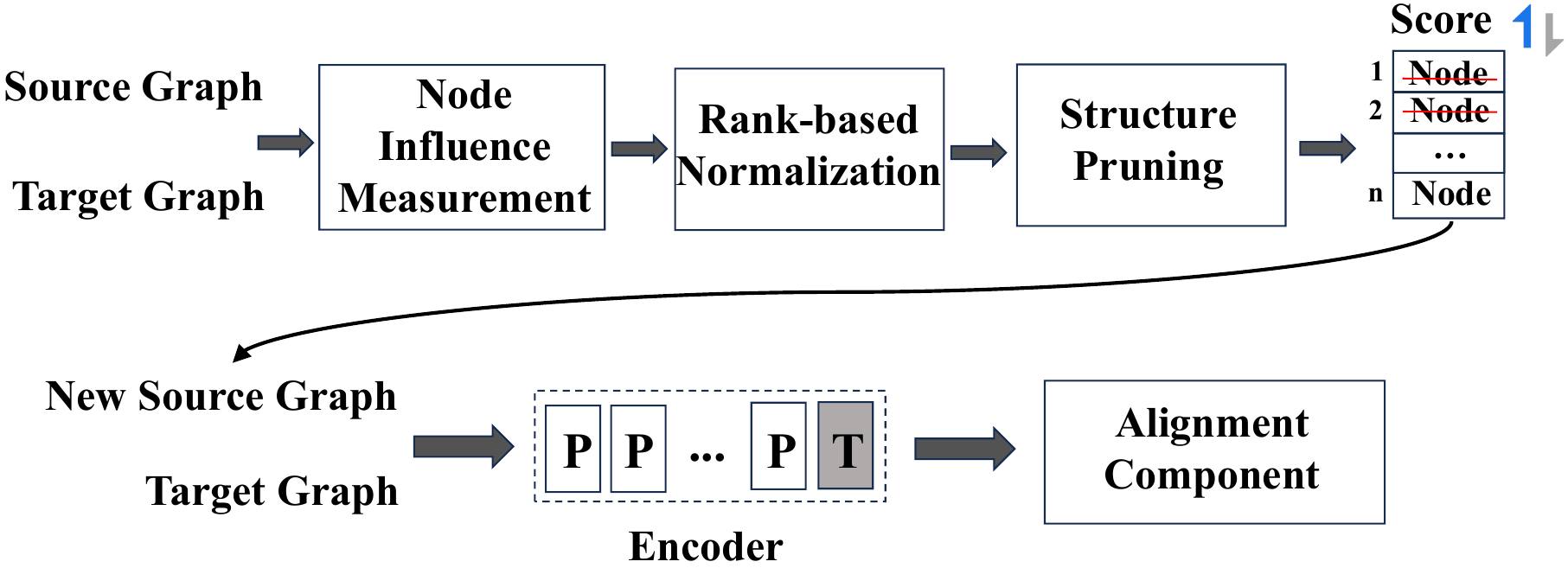}
    \caption{The overall pipeline of Source Node Influence Pruning (SNIP) framework. It filters out structurally incompatible source nodes through three cascaded steps, i.e., Node Influence Measurement, Rank-based Normalization, and Structure Pruning, and then feeds the refined source graph into the encoder and alignment component for transferable feature learning.}
    \label{fig:architecture-snip}
\end{figure}

The aforementioned empirical analysis presents the significant cross-domain distributional shifts and highlight the motivation and necessity of selective source node utilization. Building upon these insights, we propose \textbf{S}ource \textbf{N}ode \textbf{I}nfluence \textbf{P}runing (\textbf{SNIP}), a lightweight, data-centric pruning strategy explicitly designed to filter out source nodes that are structurally incompatible with the target domain.
By discarding these noisy nodes, SNIP prevents negative knowledge transfer effectively.
As illustrated in Fig.~\ref{fig:architecture-snip}, SNIP comprises three stages: 
(1) \textit{Node Influence Measurement}, which quantifies the structural discrepancy between individual source nodes and the target domain to assign node influence scores, using multi-dimensional centrality measures; 
(2) \textit{Rank-based Normalization}, which is a robust normalization mechanism to mitigate scale variation across multiple centrality measures; 
(3) \textit{Structure Pruning}, which dynamically removes outlier nodes to mitigate negative knowledge transfer.
The refined source graph is then utilized to train the GNN encoder, ensuring the model focuses on transferable structural patterns.
The following parts detail the above three stages, along with specific implementation and optimization settings.

\vspace{1ex}
\noindent
\textbf{Node Influence Measurement.} 
To filter out source nodes that are structurally incompatible with the target domain, we employ node influence scores as the primary pruning criterion. 
Consequently, the accurate measurement of node influence is critical. 
To this end, we characterize node topological information via four complementary perspectives of centrality: \textit{Degree}($d$), \textit{Betweenness}($b$), \textit{Closeness}($c$), and \textit{Eigenvector}($e$). 
By integrating these measures, we are able to capture diverse structural attributes, ranging from local connectivity and routing bottlenecks to global accessibility and influential neighborhoods.
Since these measures represent well-established foundational concepts in graph theory and complex network analysis, we omit their standard definitions here to maintain brevity. 
Further details are provided in Section~\ref{sec:experimental-settings}.

Given source graph $\mathcal{G}^s=(\mathcal{V}^s,\mathcal{E}^s)$ and target graph $\mathcal{G}^t=(\mathcal{V}^t,\mathcal{E}^t)$, we first compute the four centrality metrics for all nodes of both graphs. Subsequently, we derive the global mean value for each centrality $i \in \{d,b,c,e\}$ within the target domain $\mathcal{G}^t$.
Let $\bar{C}_i^t$ denotes this global mean, defined as:
\begin{equation}
    \bar{C}_i^t = \frac{1}{|\mathcal{V}^t|} \sum_{v^t \in \mathcal{V}^t} C_i^t(v^t),
\end{equation}
\noindent
where $C_i^t(v^t)$ denotes the value of centrality $i$ for target node $v^t$, and $|\mathcal{V}^t|$ is the number of target nodes. 
Then, we quantify the cross-domain structural discrepancy $\Delta C_i(v)$ for each source node $v \in \mathcal{V}^s$, defined as:
\begin{equation}
    \label{eq:c_i_s_v}
    \Delta C_i^s(v) = |C_i^s(v) - \bar{C}_i^t|,
\end{equation}
\noindent
where $C_i^s(v)$ denotes the value of centrality $i$ for source node $v$, and ``$|\cdot|$'' represents taking the absolute value.

Intuitively, the higher the value of structural discrepancy $\Delta C_i^s(v)$ for source node $v$, the lower its node influence.
Source nodes with high $\Delta C_i^s$ values may be identified as structural outliers incompatible with the target domain's topology, which are more likely to induce negative transfer, as their neighborhood structures differ fundamentally from those in the target domain.
Therefore, the node influence score of centrality $i$ for source node $v$ can be defined as:
\begin{equation}
    \label{eq:influence-score-definition}
    S^s_i(v) = \frac{1}{\Delta C_i^s(v)},
\end{equation}
Although computing certain global measures (e.g., betweenness centrality) may be time-consuming in the worst case, they are executed only once as a one-off offline pre-processing step, introducing zero overhead to the training phase. Furthermore, the indicators are flexible, for large-scale graphs, computationally intensive indicators can be seamlessly replaced with highly efficient, linear-time measures to guarantee scalability.

\vspace{1ex}
\noindent
\textbf{Rank-based Normalization.} 
Directly aggregating four raw node influence scores is problematic due to the vastly different scales of centrality metrics (e.g., degree is often an integer while betweenness is a small fraction). To ensure numerical robustness, we employ a Rank-based Normalization (RN) mechanism. 
For each centrality type $i$, all source nodes are first sorted by ascending order based on their difference node influence scores $S_i^s(v)$. 
Then, we can map these node influence scores into a normalized interval $[0, 1]$, defined as:
\begin{equation}
    \hat{R}_i(v) = \frac{R_i(v) - 1}{|\mathcal{V}^s| - 1},
    \label{eq:rank-norm}
\end{equation}
\noindent
where $R_i(v) \in \{1, \dots, |\mathcal{V}^s|\}$ denotes the rank of node $v$ in terms of centrality $i$, and $\hat{R}_i(v)$ can serve as the new node influence score of centrality $i$ for source node $v$.

\vspace{1ex}
\noindent\textbf{Structure Pruning.} 
To comprehensively utilize the above four centrality metrics, the final node influence score is defined as the weighted sum of all normalized centrality ranks:
\begin{equation}
    S(v) = \sum_{i \in \{d,b,c,e\}} w_i \cdot \hat{R}_i(v),
    \label{eq:sum-centrality}
\end{equation}
\noindent
where $w_i$ is the weight coefficient for centrality $i$. 
A higher $S(v)$ indicates that node $v$ exhibits a topological pattern more similar to the target domain.  
All nodes in $\mathcal{V}^s$ are sorted in descending order of $S(v)$ scores. 
According to a predefined pruning ratio $r$, the last $\lfloor r \cdot |\mathcal{V}^s| \rfloor$ nodes in the sorted sequence are removed from the source graph.
This adaptive pruning process refines the source domain by retaining only those nodes that are structurally compatible with the target domain, thereby facilitating more effective domain adaptation. The overall process of SNIP is shown in Algorithm~\ref{alg:algorithm}.

\begin{algorithm}[!tbp]
    \caption{Source Node Influence Pruning}
    \label{alg:algorithm}
    \textbf{Input}: Source graph $\mathcal{G}^s = (\mathcal{V}^s, \mathcal{E}^s)$, Target graph $\mathcal{G}^t = (\mathcal{V}^t, \mathcal{E}^t)$, weight coefficients $\{w_d, w_b, w_c, w_e\}$ for different centrality metrics, and pruning ratio $r$.\\
    \textbf{Output}: Set of nodes to be removed $\mathcal{V}_{\mathrm{prune}}$. 
    
    \begin{algorithmic}[1] %[1] enables line numbers
        \STATE For source graph $\mathcal{G}^s$, compute node-level centrality scores: Degree $C_d^s(v)$, Betweenness $C_b^s(v)$, Closeness $C_c^s(v)$, and Eigenvector $C_e^s(v)$ for each node $v \in \mathcal{V}^s$. 
        
        \STATE For target graph $\mathcal{G}^t$, compute the global mean centrality values: $\bar{C}_d^t, \bar{C}_b^t, \bar{C}_c^t, \bar{C}_e^t$.
        
        \STATE For each node $v \in \mathcal{V}^s$ and each metric $i \in \{d, b, c, e\}$, calculate the structural discrepancy from the target domain: $\Delta C_i^s(v) = |C_i^s(v) - \bar{C}_i^t|$, deriving the node influence score $S^s_i(v) = \frac{1}{\Delta C_i^s(v)}$.
        \FOR{each metric $i \in \{d, b, c, e\}$}
        \STATE Generate ranks $R_i(v)$ for all $S^s_i(v)$ in ascending order.
        
        \STATE Normalize the ranks to the range $[0, 1]$: \\$\hat{R}_i(v) = \frac{R_i(v) - 1}{|\mathcal{V}^s| - 1}$.
        
        \ENDFOR
        \STATE For each node $v \in \mathcal{V}^s$, calculate the final node influence score $S(v)$ using the predefined weights: \\
        $S(v) = w_d \cdot \hat{R}_d(v)+w_b \cdot \hat{R}_b(v)+w_c \cdot \hat{R}_c(v)+w_e \cdot \hat{R}_e(v)$.
        
        \STATE Sort all nodes $v \in \mathcal{V}^s$ in descending order based on $S(v)$.
        \STATE Determine the number of source nodes to be pruned: $N_{\mathrm{remove}} = \lfloor r \cdot |\mathcal{V}^s| \rfloor$.
        \STATE Identify $\mathcal{V}_{\mathrm{prune}}$ as the last $N_{\mathrm{remove}}$ nodes in the sorted list (nodes with the lowest influence scores).
        \STATE \textbf{return} $\mathcal{V}_{\mathrm{prune}}$.
    \end{algorithmic}
\end{algorithm}

Since our SNIP operates only at the data level, it remains entirely model-agnostic, and can serve as a plug-and-play module for any existing baseline models, such as, AdaGCN \cite{Dai2023Graph}, A2GNN \cite{Liu2024Rethinking}, etc.
Taking the A2GNN backbone as an example, the overall objective based on Maximum Mean Discrepancy (MMD) \cite{gretton2012a-kernel} can be formulated  as follows:
\begin{equation}
    \mathcal{L}_{\mathrm{total}} = \mathcal{L}_{\mathrm{cls}} + \alpha \cdot \mathcal{L}_{\mathrm{mmd}},
\end{equation}
\noindent
where $\mathcal{L}_{\mathrm{cls}}$ represents the cross-entropy loss for node classification on source graph, and $\mathcal{L}_{\mathrm{mmd}}$ focuses on minimizing the distribution discrepancy between source and target features. 
The hyper-parameter $\alpha$ balances the trade-off between classification accuracy and domain alignment.

\subsection{Theoretical Analysis}
In this subsection, we theoretically analyze how the Source Node Influence Pruning strategy affects the target-domain generalization bound. 
We first revisit the graph domain adaptation bound adopted in prior studies~\cite{shen2018wasserstein,You2023Graph}. 
We then compare the complete target-risk upper bounds before and after source-node pruning and derive a sufficient condition under which SNIP strictly tightens the risk bound.

\vspace{1ex}
\noindent
\textbf{Theorem 1.}
Let $f$ and $c$ denote the feature extractor and classifier, respectively, and let $h=c \circ f \in \mathcal{H}$ be a graph neural network hypothesis, where $\mathcal{H}$ is the hypothesis space of GNNs. Suppose that $f$ and $c$ are $K_f$- and $K_c$-Lipschitz continuous, where $K_f$ and $K_c$ are Lipschitz constants. Then, with probability at least $1-\delta$, the target domain risk $\epsilon_t(h, \hat{h})$ satisfies the following inequality \cite{shen2018wasserstein,You2023Graph}:
\begin{equation}
\begin{aligned}
  \epsilon_t(h,\hat{h}) \leq & \ \widehat{\epsilon}_s(h,\hat{h}) + 2K_fK_c W_1\big(\mathbb{P}_s, \mathbb{P}_t \big) \\
  & + \mathcal{R}\left(n_s, d, \delta\right) + \omega,
\end{aligned}
\end{equation}
\noindent
where $\widehat{\epsilon}_s(h,\hat{h})$ denotes the empirical source-domain risk, 
$\hat{h}$ is the labeling function,
$W_1(\mathbb{P}_s, \mathbb{P}_t)$ is the 1-Wasserstein distance between source- and target domain distributions, and
\begin{equation}
   \mathcal{R}(n_s, d, \delta) = \sqrt{\frac{4d}{n_s} \log\left(\frac{e n_s}{d}\right) + \frac{1}{n_s} \log\left(\frac{1}{\delta}\right)}
\end{equation}
is the finite-sample complexity term. Here $n_s$ is the number of source nodes, $d$ is the pseudo-dimension of the hypothesis, 
and $\omega$ is the joint error of the ideal hypothesis.

Theorem 1 indicates that reducing the source-target discrepancy (i.e., $W_1(\mathbb{P}_s, \mathbb{P}_t)$) can tighten the target-risk upper bound. 
However, discrepancy reduction alone is insufficient to establish the theoretical benefit of our Source Node Influence Pruning. 
Pruning also reduces the effective source sample size and may change the empirical source risk, the complexity term and the ideal joint error. 
Therefore, the complete upper bounds before and after pruning must be compared.

By filtering out $k$ structural anomalies $\mathcal{V}_{\mathrm{prune}}\subseteq\mathcal{V}^s$, SNIP yields a sub-source graph $\mathcal{G}^s_{\mathrm{sub}} = \mathcal{G}^s \setminus \mathcal{V}_{\mathrm{prune}}$ with $n_s^{\prime}=n_s-k$ retained nodes. 
Let $\mathbb{P}_s^{\prime}$ denote the corresponding distribution induced by $\mathcal{G}^s_{\mathrm{sub}}$. 
Applying Theorem~1 to the full source graph gives the upper bound
\begin{equation}
\label{eq:full-bound}
\begin{aligned}
\mathcal{B}_{\mathrm{full}} = \; & \widehat{\epsilon}_s(h_{\mathrm{full}}, \hat{h}) + 2K_f^{\mathrm{full}}K_c^{\mathrm{full}} W_1(\mathbb{P}_s,\mathbb{P}_t) \\
&+ \mathcal{R}(n_s,d,\delta) + \omega_{\mathrm{full}},
\end{aligned}
\end{equation}
whereas applying it to the retained sub-source graph gives 
\begin{equation}
\label{eq:snip-bound}
\begin{aligned}
\mathcal{B}_{\mathrm{sub}} = \; & \widehat{\epsilon}_s^{\prime}(h_{\mathrm{sub}}, \hat{h})+2K_f^{\mathrm{sub}}K_c^{\mathrm{sub}} W_1(\mathbb{P}_s^{\prime},\mathbb{P}_t)\\
&+ \mathcal{R}(n_s^{\prime},d,\delta) + \omega_{\mathrm{sub}}.
\end{aligned}
\end{equation}

We define the reduction in the discrepancy-related term as the \textit{alignemnt gain}:
\begin{equation}
\label{eq:alignment-gain}
\begin{aligned}
\mathcal{T}_{\mathrm{align}} = 2\Big[ &K_f^{\mathrm{full}}K_c^{\mathrm{full}} W_1(\mathbb{P}_s,\mathbb{P}_t)\\
&-K_f^{\mathrm{sub}}K_c^{\mathrm{sub}}W_1(\mathbb{P}_s^{\prime},\mathbb{P}_t)\Big].
\end{aligned}
\end{equation}
The remaining changes are defined as the \textit{pruning cost} term:
\begin{equation}
\label{eq:pruning-cost}
\begin{aligned}
\mathcal{T}_{\mathrm{prune}} = \;
& \widehat{\epsilon}_s^{\prime}(h_{\mathrm{sub}},\hat{h}) - \widehat{\epsilon}_s(h_{\mathrm{full}}, \hat{h}) \\
& +\mathcal{R}(n_s^{\prime},d,\delta) - \mathcal{R}(n_s,d,\delta) \\
& +\omega_{\mathrm{sub}} - \omega_{\mathrm{full}}.
\end{aligned}
\end{equation}
The pruning cost accounts for the changes in empirical source risk,
finite-sample complexity, and ideal joint error. In particular,
because $n_s'<n_s$, pruning generally increases the finite-sample
complexity term $\mathcal{R}(\cdot,d,\delta)$.

\vspace{1ex}
\noindent
\textbf{Proposition 1.}
Our Source Node Influence Pruning strategy strictly tightens the target-risk upper bound if
\begin{equation}
\label{eq:sufficient-condition}
\mathcal{T}_{\mathrm{align}} > \mathcal{T}_{\mathrm{prune}}.
\end{equation}

\vspace{1ex}
\noindent
\textbf{Proof.}
Subtracting Eq.~\eqref{eq:full-bound} from
Eq.~\eqref{eq:snip-bound} yields
\begin{equation}
\mathcal{B}_{\mathrm{sub}} - \mathcal{B}_{\mathrm{full}}
= \mathcal{T}_{\mathrm{prune}} - \mathcal{T}_{\mathrm{align}}.
\end{equation}
Therefore,
$\mathcal{T}_{\mathrm{align}}>\mathcal{T}_{\mathrm{prune}}$
implies
$\mathcal{B}_{\mathrm{sub}}<\mathcal{B}_{\mathrm{full}}$.
\hfill $\square$

\vspace{1ex}
Proposition~1 shows that our source-node pruning strategy does not unconditionally improve the generalization bound. 
Instead, the alignment gain induced by pruning must compensate for the finite-sample complexity penalty and the potential changes in the empirical source risk and ideal joint error.

We next relate this sufficient condition in Proposition~1 to the node-selection mechanism of SNIP. 
According to Eqs.~\ref{eq:c_i_s_v} and~\ref{eq:influence-score-definition} in Section~\ref{sec:source-node-influence-pruning},
the influence score $S_i^s(v)$ of each source node $v$ is defined as a monotonically decreasing function of its cross-domain structural discrepancy $\Delta C_i(v)$.
Consequently, source nodes that are structurally closer to the target domain receive higher influence scores, whereas source nodes with larger structural discrepancies receive lower scores.

Our SNIP ranks the source nodes in descending order of their influence scores and prunes a predefined proportion of nodes from the end of the ranking. 
Since the influence score decreases monotonically with the structural discrepancy, this procedure is equivalent to preferentially pruning source nodes with the largest structural discrepancies across domains,
Therefore, under a fixed pruning ratio, SNIP is designed to
promote a positive discrepancy reduction and thereby increase the alignment gain $\mathcal{T}_{\mathrm{align}}$ as much as possible.

For a simplified interpretation, suppose that the models before and
after pruning share the same Lipschitz constant upper bounds
$\overline{K}_f$ and $\overline{K}_c$, and that pruning introduces
negligible changes in the empirical source risk and the ideal joint
error. A sufficient condition to tighten the target-risk upper bound is 
\begin{equation}
\label{eq:simplified-condition}
\begin{aligned}
2\overline{K}_f\overline{K}_c
\big[W_1(\mathbb{P}_s,\mathbb{P}_t)&-W_1(\mathbb{P}_s^{\prime},\mathbb{P}_t)\big]\\
&>
\mathcal{R}(n_s^{\prime},d,\delta) - \mathcal{R}(n_s,d,\delta).
\end{aligned}
\end{equation}

This provides a direct interpretation of our SNIP strategy: source-node pruning is beneficial when the
reduction in the source-target discrepancy outweighs the finite-sample
penalty caused by the reduced source node size. 
This result also explains why an appropriate pruning ratio is necessary.
For detailed results, see Fig.~\ref{fig:empirical-observation-2} in Section~\ref{sec:empirical-analysis} and Fig.~\ref{fig:sensitivity_analysis_pruning_ratio} in Section~\ref{sec:sensitivity-analysis}.
Insufficient pruning may retain structurally incompatible source nodes, whereas excessive pruning may discard informative nodes and introduce a
pruning cost that exceeds the attainable alignment gain.

\section{Experiments and Results}
In this section, we first outline the experimental settings, followed by a comprehensive comparison between our method and all baseline models on the cross-network node classification task.
Next, we conduct an ablation study to investigate the contribution of each component within our SNIP framework, along with sensitivity analyses regarding its key hyper-parameters.
Lastly, we present t-SNE visualizations to illustrate the representations learned by our SNIP method.

\subsection{Experimental Settings}
\label{sec:experimental-settings}

\noindent
\textbf{Datasets.} 
We evaluate the performance of our SNIP method on five real-world datasets, including three citation networks from ArnetMiner~\cite{Tang2008ArnetMiner}\footnote{https://github.com/yuntaodu/ASN/tree/main/data}: ACMv9 (A), Citationv1 (C), and DBLPv7 (D), alongside two social networks from the Twitch gamer networks~\cite{rozemberczki2021multi}\footnote{https://snap.stanford.edu/data/twitch-social-networks.html}: Germany (DE) and England (EN). 
In the citation datasets, nodes represent academic papers and edges indicate citation relationships, whereas in the social networks, nodes and edges constitute users and their social connections, respectively.
The detailed statistical properties of these datasets are summarized in Table~\ref{tab:table-dataset-statistics}.
Specifically, we construct six cross-domain adaptation scenarios on the citation networks, including A$\rightarrow$C, A$\rightarrow$D, C$\rightarrow$A, C$\rightarrow$D, D$\rightarrow$A, and D$\rightarrow$C, as well as two cross-domain adaptation scenarios on the social networks: DE$\rightarrow$EN and EN$\rightarrow$DE.

\begin{table}[!tbp]
\caption{Detailed statistical properties of the citation and social network datasets.}
\label{tab:table-dataset-statistics}
\centering
\resizebox{\linewidth}{!}{%
\begin{tabular}{l|l|c|c|c|c}
\toprule
Types & Datasets & \#Node & \#Edge & \#Feat & \#Label \\
\specialrule{0.5pt}{3pt}{3pt}
 & ACMv9 & 9,360 & 15,556 & \multirow{3}{*}{6,775} & \multirow{3}{*}{5} \\
Citation Networks & Citationv1 & 8,935 & 15,098 &  &  \\
 & DBLPv7 & 5,484 & 8,117 &  &  \\
\specialrule{0.5pt}{3pt}{3pt}
\multirow{2}{*}{Social Networks} & Germany & 9,498 & 153,138 & \multirow{2}{*}{3,170} & \multirow{2}{*}{2} \\
 & England & 7,126 & 35,324 &  &  \\
\bottomrule
\end{tabular}
}
\end{table}

\vspace{1ex}
\noindent
\textbf{Baselines.} 
To comprehensively evaluate the performance of our SNIP framework, we compare it against a wide range of state-of-the-art baselines, which can be broadly categorized into three representative groups:

(1) Hypothesis transfer methods based on unsupervised graph representation learning, including DeepWalk \cite{Perozzi2014DeepWalk}, node2vec \cite{Grover2016node2vec}, and ANRL \cite{Zhang2018ANRL}. These methods typically learn node embeddings in an unsupervised manner based solely on graph topology or node attributes, 
and directly apply the classifiers trained on the source domain to evaluate target node representations without explicitly handling domain shifts;

(2) Source-only GNN-based methods, including GAT \cite{Velickovic2017Graph}, GSAGE \cite{Hamilton2017Inductive}, SGC \cite{Wu2019Simplifying}, GIN~\cite{xu2019how} and GCN \cite{Kipf2017Semi}. These models are trained in a supervised manner on the source graph and then directly applied to the target graph for inference;

(3) Graph domain adaptation methods, including CDNE \cite{Shen2021Network}, AdaGCN \cite{Dai2023Graph}, ACDNE \cite{Shen2020Adversarial}, UDAGCN \cite{Wu2020Unsupervised}, ASN~\cite{Zhang2021Adversarial}, GRADE \cite{Wu2023Non}, StruRW \cite{Liu2023Structural}, SpecReg \cite{You2023Graph}, PairAlign \cite{Liu2024Pairwise}, A2GNN \cite{Liu2024Rethinking}, TDSS \cite{Chen2025Smoothness}, TASSC~\cite{zou2025target}, and DAGNN~\cite{yang2026dagnn}. These methods employ adversarial training, structural rewiring, or spectral regularization to align two graphs, representing the most competitive baselines.

\vspace{1ex}
\noindent
\textbf{Implementation Details.} 
We train our proposed SNIP model using the pruned source nodes alongside all target nodes, while model evaluation is conducted exclusively on the target-domain nodes.
Specifically, SNIP is implemented based on the PyTorch framework \cite{Paszke2019PyTorch} and optimized using the Adam optimizer over a maximum of 250 epochs.
To prevent overfitting, we implement an early stopping mechanism. 
For the model architecture, we adopt AdaGCN~\cite{Dai2023Graph} and A2GNN~\cite{Liu2024Rethinking} as our default backbone networks. All of them use the official standard configurations, and A2GNN leverages a Gaussian kernel to compute the Maximum Mean Discrepancy (MMD)~\cite{gretton2012a-kernel} for cross-domain alignment.
To identify the optimal configuration, a comprehensive grid search is performed over two primary sets of hyper-parameters: the node pruning ratio $r\in\{10\%, 15\%, 20\%, 25\%\}$, and the trade-off coefficients $\{w_d,w_b,w_c,w_e\}$ for the joint centrality metrics in Eq.~\ref{eq:sum-centrality}, which are thoroughly sampled from the list $[0,0.1,0.2,0.3,0.4,0.5,0.6,0.7,0.8,0.9,1.0]$.
All subsequent experiments are deployed on a single NVIDIA GeForce RTX 3090 GPU. We report both Micro-F1 and Macro-F1 scores, averaged over three independent runs with distinct random seeds to ensure statistical stability. 

\vspace{1ex}
\noindent
\textbf{Definitions of Influence Measurement Indicators.}
Since our SNIP framework computes node influence scores based on multiple structural indicators, we briefly introduce the candidate structural indicators considered in this paper below:

\begin{itemize}
    \item Degree Centrality indicates the direct connectivity of a node in the graph, measured by the number of edges directly connected to the node.
    \item Betweenness Centrality indicates the extent to which a node serves as a bridge or intermediary between other nodes, measured by the number or proportion of shortest paths between node pairs that pass through the node.
    \item Closeness Centrality indicates how efficiently a node can reach all other nodes in the graph, measured by the reciprocal of the average shortest-path distance from the node to all other reachable nodes.
    \item Eigenvector Centrality indicates the influence of a node by considering not only its connections but also the importance of its connected neighbors, measured by the corresponding value of the node in the principal eigenvector of the adjacency matrix.
    \item PageRank indicates the global importance of a node under a random-walk-based propagation process, measured by the probability that a random walker visits the node.
    \item Clustering Coefficient indicates the local clustering or cohesiveness around a node, measured by the ratio of existing edges among the node’s neighbors to the total number of possible edges among those neighbors.
    \item $K$-Core Number indicates the structural coreness of a node within the graph, measured by the largest integer $K$ such that the node belongs to a $K$-core subgraph.
    \item Average Neighbor Degree indicates the average connectivity level and structural density of a node’s local neighborhood, measured by the mean degree of all neighboring nodes that are directly connected to the node.
\end{itemize}

\begin{table*}[!t]
\caption{Overall performance (Micro-F1 / Macro-F1) of node classification task across six transfer scenarios on citation networks.}
\label{tab:main}
\centering
% \label{tab:cross-domain}
\resizebox{\linewidth}{!}{%
\begin{tabular}{l|cc|cc|cc|cc|cc|cc|cc}
\toprule
\multirow{2}{*}{Methods} & \multicolumn{2}{c|}{A$\rightarrow$C} & \multicolumn{2}{c|}{A$\rightarrow$D} & \multicolumn{2}{c|}{C$\rightarrow$A} & \multicolumn{2}{c|}{C$\rightarrow$D} & \multicolumn{2}{c|}{D$\rightarrow$A} & \multicolumn{2}{c|}{D$\rightarrow$C} & \multicolumn{2}{c}{Avg.}\\
\cline{2-15}
 & Micro & Macro & Micro & Macro & Micro & Macro & Micro & Macro & Micro & Macro & Micro & Macro & Micro & Macro \\
\specialrule{0.5pt}{3pt}{3pt}

DeepWalk & {21.05} & {17.72} & {25.94} & {19.87} & {21.94} & {19.33} & {22.57} & {17.51} & {26.23} & {19.83} & {29.46} & {22.76} & {24.53} & {19.56} \\
node2vec & {29.89} & {25.84} & {24.54} & {19.50} & {21.76} & {17.99} & {28.95} & {24.98} & {28.61} & {22.05} & {21.16} & {16.22} & {25.82} & {21.08} \\
ANRL & {30.31} & {20.93} & {29.54} & {23.33} & {31.84} & {22.04} & {25.90} & {22.71} & {29.56} & {19.12} & {25.99} & {18.25} & {28.86} & {20.99} \\

GAT & {57.13} & {43.64} & {53.80} & {41.36} & {50.37} & {42.14} & {55.85} & {45.25} & {52.93} & {43.95} & {55.52} & {50.04} & {54.27} & {44.36} \\
GSAGE & {71.40} & {69.14} & {64.82} & {61.80} & {65.22} & {64.69} & {69.96} & {66.86} & {59.22} & {57.31} & {67.85} & {64.90} & {66.41} & {63.79} \\
SGC & {77.40} & {72.13} & {69.13} & {62.54} & {72.31} & {62.92} & {73.80} & {66.93} & {63.32} & {53.94} & {72.31} & {62.92} & {71.38} & {62.90} \\
GCN & {77.38} & {74.78} & {69.05} & {65.29} & {70.58} & {70.39} & {74.17} & {71.37} & {63.35} & {59.42} & {74.17} & {69.79} & {71.45} & {68.24} \\

UDAGCN & {72.15} & {60.33} & {66.95} & {64.83} & {66.80} & {67.22} & {71.77} & {69.46} & {58.16} & {55.89} & {73.28} & {61.12} & {68.18} & {63.14} \\
GRADE & {76.04} & {72.52} & {68.22} & {63.03} & {69.55} & {69.34} & {73.95} & {70.02} & {63.72} & {59.35} & {74.32} & {69.32} & {70.96} & {67.26} \\
CDNE & {78.76} & {76.83} & {71.58} & {69.24} & {74.22} & {75.06} & {74.36} & {71.34} & {69.62} & {70.45} & {78.88} & {77.36} & {74.57} & {73.38} \\
% AdaGCN & {79.32} & {76.51} & {75.04} & {71.39} & {71.67} & {70.77} & {75.59} & {72.34} & {69.67} & {69.47} & {78.20} & {74.22} & {74.91} & {72.45} \\
StruRW & {77.35} & {72.07} & {69.10} & {62.51} & {67.81} & {59.77} & {73.81} & {66.89} & {63.27} & {53.82} & {72.41} & {62.94} & {70.16} & {62.34} \\
PairAlign & {70.88} & {67.88} & {65.91} & {62.35} & {65.85} & {65.09} & {71.04} & {67.56} & {59.34} & {58.77} & {67.07} & {64.61} & {66.02} & {63.38}\\
SpecReg & {80.55} & {78.83} & {75.93} & {73.98} & {72.04} & {73.15} & {75.74} & {73.64} & {71.01} & {72.34} & {79.04} & {77.78} & {75.71} & {74.95} \\
ACDNE & {81.75} & {80.09} & {76.24} & {73.59} & {73.59} & {74.79} & {77.21} & {75.74} & {71.29} & {72.64} & {80.14} & {78.83} & {76.70} & {75.94} \\
TDSS & {83.20} & {81.40} & {79.54} & {78.13} & {76.94} & {78.33} & {78.96} & {77.55} & {74.89} & {76.46} & {82.29} & {80.75} & {79.54} & {79.06} \\
TASSC & 83.95 & 82.94 & 77.91 & 76.11 & 75.41 & 77.11 & 78.40 & 77.23 & 74.16 & 75.68 & 83.19 & 82.11 & 78.84 & 78.53 \\
DAGNN & 83.41 & 81.65 & 79.06 & 76.87 & 76.32 & 78.15 & 78.93 & 77.49 & 74.81 & 76.24  & 82.24 & 80.17 & 79.13 & 78.43\\

\specialrule{0.5pt}{3pt}{3pt}
AdaGCN & 79.32 & 76.51 & 75.04 & 71.39 & 69.67 & 69.47 & 75.59 & 72.34 & 69.67 & 69.47 & 78.20 & 74.22  & 74.91 & 72.45\\
\textbf{SNIP}$_{\text{AdaGCN}}$ & \textbf{83.03} & \textbf{81.16} & \textbf{76.86} & \textbf{75.05} & \textbf{76.60} & \textbf{77.26} & \textbf{76.74} & \textbf{75.34} & \textbf{73.59} & \textbf{74.06} & \textbf{80.67} & \textbf{79.55} & \textbf{77.92} & \textbf{77.07} \\
Improv.(\%) & +3.71 & +4.65 & +1.82 & +3.66 & +6.93 & +7.79 & +1.15 & +3.00 & +3.92 & +4.59 & +2.47 & +5.33 & +3.01 & +4.62 \\

\specialrule{0.5pt}{3pt}{3pt}
A2GNN & {82.75} & {80.86} & {77.84} & {75.18} & {74.63} & {76.31} & {78.28} & {75.90} & {73.44} & {74.94} & {81.09} & {78.84} & {77.95} & {76.72} \\
\textbf{SNIP}$_{\text{A2GNN}}$ & \textbf{83.13} & \textbf{82.09} & \textbf{79.10} & \textbf{78.32} & \textbf{77.05} & \textbf{78.35} & \textbf{79.27} & \textbf{78.05} & \textbf{75.11} & \textbf{76.71} & \textbf{82.00} & \textbf{80.18} & \textbf{79.27} & \textbf{78.95}\\
Improv.(\%) & {+0.38} & {+1.23} & {+1.26} & {+3.14} & {+2.42} & {+2.04} & {+0.99} & {+2.15} & {+1.67} & {+1.77} & {+0.91} & {+1.34} & {+1.32} & {+2.23} \\
\bottomrule
\end{tabular}
}
\end{table*}

\begin{table}[!tbp]
\caption{Overall performance (Micro-F1 / Macro-F1) of node classification task on social networks.}
\label{tab:main2}
\centering
% \resizebox{\linewidth}{!}{%
\begin{tabular}{l|cc|cc|cc}
\toprule
\multirow{2}{*}{Methods} & \multicolumn{2}{c|}{DE$\rightarrow$EN} & \multicolumn{2}{c|}{EN$\rightarrow$DE} & \multicolumn{2}{c}{Avg.} \\
\cline{2-7}
& Micro & Macro & Micro & Macro & Micro & Macro\\
\specialrule{0.5pt}{3pt}{3pt}
node2vec & 52.64 & 46.96 & 54.61 & 50.10 & 53.62 & 48.53\\
DeepWalk & 52.18 & 46.54 & 55.08 & 49.97 & 53.63 & 48.25\\
GAT & 54.84 & 49.50 & 43.65 & 40.08 & 49.24 & 44.79\\
GCN & 54.77 & 54.55 & 62.03 & 51.04 & 58.40 & 52.79\\
GIN & 52.39 & 49.91 & 55.26 & 44.26 & 53.82 & 47.08\\
UDAGCN & 59.74 & 58.19 & 58.69 & 56.35 & 59.21 & 57.27\\
ACDNE & 58.08 & 56.31 & 58.79 & 57.92 & 58.43 & 57.11\\
ASN & 55.45 & 51.21 & 60.45 & 45.90 & 57.95 & 48.55\\
% AdaGCN & 54.56 & 35.30 & 40.22 & 31.18 & 47.39 & 33.24\\+
GRADE & 56.40 & 56.38 & 61.18 & 56.83 & 58.79 & 56.60\\
SpecReg & 56.43 & 50.30 & 61.45 & 46.13 & 58.94 & 48.21\\
TASSC & 58.74 & 57.08 & 64.38 & 62.25 & 61.56 & 59.67\\
DAGNN & 61.05 & 60.28 & 65.42 & 62.93 & 63.24 & 61.61\\

\specialrule{0.5pt}{3pt}{3pt}
AdaGCN & 56.03 & 54.57 & 59.99 & 58.87 & 58.01 & 56.72\\
\textbf{SNIP}$_{\text{AdaGCN}}$ & \textbf{56.65} & \textbf{55.31} & \textbf{62.70} & \textbf{60.41} & \textbf{59.67} & \textbf{57.86} \\
Improv.(\%) & +0.62 & +0.74 & +2.71 & +1.54 & +1.66 & +1.14\\

\specialrule{0.5pt}{3pt}{3pt}
A2GNN & 59.54 & 58.44 & 63.90 & 61.42 & 61.72 & 59.93\\
\textbf{SNIP}$_{\text{A2GNN}}$ & \textbf{60.67} & \textbf{59.13} & \textbf{65.21} & \textbf{63.08} & \textbf{62.94} & \textbf{61.10} \\
Improv.(\%) & +1.13 & +0.69 & +1.31 & +1.66 & +1.22 & +1.17\\
\bottomrule
\end{tabular}
% }
\end{table}

\subsection{Main Results}
We conduct experiments on two types of networks: citation networks and social networks.
The main results comparing SNIP with all baseline methods are presented in Table~\ref{tab:main} and Table~\ref{tab:main2}, respectively, where the performance of all baselines is reported by Liu et al. \cite{Liu2024Rethinking} and Chen et al. \cite{Chen2025Smoothness}.
The key observations are summarized as follows:

\textbf{(1) Incorporating SNIP consistently enhances baseline performance across diverse backbones.}
As quantified in Table~\ref{tab:main} and Table~\ref{tab:main2}, our SNIP demonstrates excellent compatibility as a plug-and-play strategy. When deployed on AdaGCN and A2GNN, SNIP delivers consistent performance improvements across all metrics. 

Specifically, under the A2GNN backbone (SNIP$_{\text{A2GNN}}$), it yields an average improvement of 2.23\% in Macro-F1 and 1.32\% in Micro-F1 across different cross-domain adaptation scenarios on citation networks in Table~\ref{tab:main}, achieving an overall average of 78.95\% for Macro-F1 and 79.27\% for Micro-F1. 
Notably, it achieves optimal or highly comparable performance against the state-of-the-art model TDSS in several scenarios, such as A$\rightarrow$D, C$\rightarrow$A and D$\rightarrow$A.
A horizontal comparison in Table~\ref{tab:main} shows that SNIP$_{\text{A2GNN}}$ ranks first in 8 out of 12 metric values and second in the remaining four, surpassed only by TDSS.
Similarly, our SNIP$_{\text{A2GNN}}$ yields an average improvement of 1.17\% in Macro-F1 and 1.22\% in Micro-F1 across different scenarios (i.e., DE$\rightarrow$EN and EN$\rightarrow$DE) on social networks in Table~\ref{tab:main2}, achieving an overall average of 61.10\% for Macro-F1 and 62.94\% for Micro-F1. The same findings are obtained for both citation networks and social networks, confirming the strong generalization performance of our proposed SNIP framework.

This robust scalability is further substantiated by its integration with AdaGCN backbone (SNIP$_{\text{AdaGCN}}$). 
As shown in Table II, SNIP$_{\text{AdaGCN}}$ triggers dramatic performance enhancements compared to the vanilla AdaGCN model, outperforming it on every evaluation metrics. On average, SNIP$_{\text{AdaGCN}}$ yields an improvement of 4.62\% in Macro-F1 and 3.01\% in Micro-F1 across six cross-domain adaptation scenarios.
The largest improvement is witnessed in the C$\rightarrow$A scenario, where SNIP$_{\text{AdaGCN}}$ brings an extraordinary gain of 7.79\% in Macro-F1 and 6.93\% in Micro-F1. 
Furthermore, this strength extends naturally into the social networks in Table~\ref{tab:main2}. Our SNIP$_{\text{AdaGCN}}$ brings an average improvement of 1.14\% in Macro-F1 and 1.66\% in Micro-F1, achieving consistent positive improvements across all evaluation metrics for the DE$\rightarrow$EN and EN$\rightarrow$DE transfer scenarios, despite the inherently complex and dense nature of social linkages. 

These consistent gains across distinct backbones and domains are attributed to the proposed source node influence pruning strategy. 
By unifying four complementary centrality metrics, our SNIP approach enables more precise regulation of individual node influence.
This design successfully filters out distributional noise and structural anomalies, thereby significantly enhancing the generalization capability and discriminative power of the learned graph representations.

\textbf{(2) Graph domain adaptation methods outperform traditional hypothesis transfer methods and source-only GNN-based methods.}
As demonstrated by Table~\ref{tab:main} and Table~\ref{tab:main2},
vanilla GNN architectures (e.g., GCN, GAT, and GIN) and traditional embedding techniques (e.g., DeepWalk and node2vec) suffer from severe performance degradation when facing cross-domain distribution shifts. 
For instance, in the citation networks (Table~\ref{tab:main}), unsupervised embedding methods (e.g., DeepWalk and node2vec) struggle heavily, yielding an average Micro-F1 of only around 25\%. 
While source-only GNN-based models capture localized neighborhood structures and perform slightly better, their generalization capabilities remain strictly constrained.
The underlying reason is that they typically presume the training (source) and testing (target) graphs share identical feature and topological distributions. 
This assumption is frequently invalidated in practical cross-domain scenarios where distinct distribution discrepancies exist, thereby leading to substantial performance degradation.
In contrast, graph domain adaptation methods are explicitly designed to mitigate such domain shifts.
By simultaneously reconciling node attribute-level shifts and graph topological discrepancies, these specialized frameworks dynamically map graph representations into a shared, domain-invariant latent subspace. Consequently, they successfully preserve the semantic discriminative power across domain boundaries.

In summary, these findings provide empirical evidence that SNIP significantly enhances the generalization capability of various GNN architectures. The consistent improvements across different scenarios and backbones highlight the effectiveness of SNIP in advancing graph domain adaptation.

\begin{table*}[!tbp]
\caption{Ablation study on cross-network node classification across six transfer scenarios within citation network datasets.}
\label{tab:ablation}
\centering
% \label{tab:cross-domain-1}
\resizebox{\linewidth}{!}{%
\begin{tabular}{l|cc|cc|cc|cc|cc|cc|cc}
\toprule
\multirow{2}{*}{Methods} & \multicolumn{2}{c|}{A$\rightarrow$C} & \multicolumn{2}{c|}{A$\rightarrow$D} & \multicolumn{2}{c|}{C$\rightarrow$A} & \multicolumn{2}{c|}{C$\rightarrow$D} & \multicolumn{2}{c|}{D$\rightarrow$A} & \multicolumn{2}{c|}{D$\rightarrow$C} & \multicolumn{2}{c}{Avg.}\\
\cline{2-15}
& Micro & Macro & Micro & Macro & Micro & Macro & Micro & Macro & Micro & Macro & Micro & Macro & Micro & Macro \\
\specialrule{0.5pt}{3pt}{3pt}
SNIP$_{\text{A2GNN}}$ & \textbf{83.13} & \textbf{82.09} & \textbf{79.10} & \textbf{78.32} & \textbf{77.05} & \textbf{78.35} & \textbf{79.27} & \textbf{78.05} & \textbf{75.11} & \textbf{76.71} & \textbf{82.00} & \textbf{80.18} & \textbf{79.27} & \textbf{78.95}\\
$\sim$ w/o. D & {82.89} & {81.99} & {77.57} & {75.92} & {76.70} & {78.23} & {79.05} & {77.33} & {72.80} & {74.07} & {81.96} & {80.11} & {78.50} & {77.94} \\
$\sim$ w/o. B & {82.94} & {81.58} & {78.28} & {76.05} & {76.37} & {77.89} & {79.21} & {77.98} & {74.96} & {76.63} & {82.04} & {80.53} & {78.97} & {78.44} \\
$\sim$ w/o. C & {82.70} & {81.69} & {77.77} & {75.02} & {76.40} & {77.99} & {78.98} & {77.43} & {74.55} & {76.30} & {81.78} & {80.04} & {78.70} & {78.07} \\
$\sim$ w/o. E & {81.30} & {79.66} & {78.45} & {76.78} & {75.40} & {76.74} & {77.70} & {76.17} & {74.66} & {76.27} & {81.82} & {79.76} & {78.22} & {77.56} \\
$\sim$ w/o. R & {80.82} & {79.08} & {77.44} & {75.46} & {74.73} & {75.85} & {73.80} & {66.93} & {73.97} & {75.00} & {81.43} & {79.47} & {77.03} & {75.29} \\
\bottomrule
\end{tabular}
}
\end{table*}

\subsection{Ablation Study}

In this section, we conduct an ablation study to assess the individual effects of each core component within SNIP, by systematically removing key components one by one.
Specifically, our SNIP based on the A2GNN backbone (SNIP$_{\text{A2GNN}}$) is compared with several derived variants, each eliminating a specific component such as a particular centrality metric or the rank-based normalization mechanism.
The results are presented in terms of Micro-F1 and Macro-F1 scores across multiple cross-domain adaptation scenarios. 
To this end, we implement and evaluate the following five ablation variants:

\begin{itemize}
    \item $\sim$ w/o. D: Degree Centrality is removed.
    \item $\sim$ w/o. B: Betweenness Centrality is removed.
    \item $\sim$ w/o. C: Closeness Centrality is removed.
    \item $\sim$ w/o. E: Eigenvector Centrality is removed.
    \item $\sim$ w/o. R: Rank-based Normalization is removed.
\end{itemize}

As illustrated in Table~\ref{tab:ablation}, the full SNIP framework consistently outperforms all of its variants on average, demonstrating the effectiveness of each module within SNIP. 
Specifically,
(1) Removing degree centrality (i.e., $\sim$ w/o. D) results in a noticeable performance drop compared to the full model, with an average reduction of 0.77\% in Micro-F1 and 1.01\% in Macro-F1.
This underscores that local connectivity attributes captured by degree distributions remain critical for identifying structurally incompatible nodes.
(2) Deleting betweenness centrality (i.e., $\sim$ w/o. B) induces a marginal performance decline, where Micro-F1 drops by 0.3\% and Macro-F1 by 0.51\%.  
This implies that although betweenness centrality is important, its absence does not significantly harm performance, suggesting that other centrality measures can somewhat compensate for the missing information. 
(3) When closeness centrality is removed (i.e., $\sim$ w/o. C), the performance also decreases, driving down Micro-F1 and Macro-F1 by 0.57\% and 0.88\%, respectively, showing the necessity of tracking global reachability within the graph topology.
(4) The variant without eigenvector centrality (i.e., $\sim$ w/o. E) leads to a notable performance decline, particularly reducing Macro-F1 by 1.39\%.
This drop indicates that removing eigenvector centrality seems to have a more pronounced effect on the model's ability to leverage global network structure. 
(5) Finally, the variant discarding the Rank-based Normalization mechanism (i.e., $\sim$ w/o. R) exhibits the most severe performance degradation, with Micro-F1 and Macro-F1 dropping substantially by 2.24\% and 3.66\%, respectively. 
This empirical evidence validates that the Rank-based Normalization plays a crucial role in performance optimization. Without it, the raw centrality values would vary vastly in scale, hindering the accurate measurement of node influence, disrupting the subsequent pruning process, and causing a failure in cross-domain alignment.

Based on the above analysis, this ablation study demonstrates that each designed component, 
% including degree centrality, betweenness centrality, closeness centrality, eigenvector centrality, and Rank-based Normalization mechanism, 
contributes to the success of cross-domain node classification task. 
Among these, the Rank-based Normalization mechanism and eigenvector centrality have the most pronounced impact on cross-domain adaptation performance, whereas betweenness centrality has a comparatively smaller effect.

\subsection{Sensitivity analysis}
\label{sec:sensitivity-analysis}

In this section, we investigate the sensitivity of our model with respect to the selection of structural indicators (e.g., various centrality measures) for calculating node influence, the trade-off coefficient $\alpha$, and the pruning ratio $r$.

\vspace{1ex}
\noindent
\textbf{Impact of Structural Indicator Selection}. 
In our SNIP framework, the quantification of node influence scores inherently depends on the combination of selected structural indicators. 
To justify our selection rationale, we systematically compare two distinct configurations using A2GNN as the backbone. 
The default configuration combines degree centrality (D), betweenness centrality (B), closeness centrality (C), and eigenvector centrality (E). 
Conversely, the alternative configuration includes PageRank (PR), Clustering Coefficient (CC), K-Core Number (KC), and Average Neighbor Degree (AD), which primarily emphasize core-periphery structure, localized clustering or specific random-walk dynamics.
Specifically, PageRank measures node importance through localized random-walk-based propagation; 
the Clustering Coefficient reflects how tightly neighbor nodes are connected; 
the K-core number characterizes a node’s position in the graph’s core–periphery hierarchy;
and the Average Neighbor Degree measures the mean connectivity density of a node's immediate neighborhood. 
Further details are provided in Section~\ref{sec:experimental-settings}.

As reported in Table~\ref{tab:structural_indicator_comparison}, the empirical results across six cross-domain transfer scenarios validate the superiority of our default indicator configuration.
Specifically, the default configuration yields the highest average classification performance, delivering a Micro-F1 of 79.27\% and a Macro-F1 of 78.95\%, which outperforms the alternative configuration by margins of $0.18\%$ and $0.43\%$, respectively. 

Crucially, it is worth highlighting that the selection of structural indicators within our SNIP framework is intrinsically decoupled and flexible, rather than rigidly fixed. 
The alternative configuration achieves superior scores in certain metrics (such as the Micro-F1 of  A$\rightarrow$C and D$\rightarrow$A), 
indicating that indicator substitution is feasible and demonstrating the high modularity of our SNIP approach. 
Since any custom set of graph structural indicators can be seamlessly integrated without altering the core components, our SNIP framework exhibits excellent architectural extensibility. 
Consequently, when dealing with large-scale graphs where the calculation of global centrality metrics (e.g., betweenness) can be very time-consuming, 
researchers can flexibly deploy highly parallelizable or localized linear-time metrics (e.g., Average Neighbor Degree or PageRank approximations) as alternatives, which can still deliver competitive performance. 
This plug-and-play property successfully ensures the scalability of our approach.

\begin{table*}[!tbp]
\caption{Performance comparison of different structural indicator configurations across six cross-domain transfer scenarios.}
\label{tab:structural_indicator_comparison}
\centering
\resizebox{\linewidth}{!}{%
\begin{tabular}{l|cc|cc|cc|cc|cc|cc|cc}
\toprule
\multirow{2}{*}{Methods} & \multicolumn{2}{c|}{A$\rightarrow$C} & \multicolumn{2}{c|}{A$\rightarrow$D} & \multicolumn{2}{c|}{C$\rightarrow$A} & \multicolumn{2}{c|}{C$\rightarrow$D} & \multicolumn{2}{c|}{D$\rightarrow$A} & \multicolumn{2}{c|}{D$\rightarrow$C} & \multicolumn{2}{c}{Avg.}\\
\cline{2-15}
& Micro & Macro & Micro & Macro & Micro & Macro & Micro & Macro & Micro & Macro & Micro & Macro & Micro & Macro \\
\specialrule{0.5pt}{3pt}{3pt}
A2GNN (Backbone) & 82.75 & 80.86 & 77.84 & 75.18 & 74.63 & 76.31 & 78.28 & 75.90 & 73.44 & 74.94 & 81.09 & 78.84 & 77.95 & 76.72 \\
Default (D, B, C, E) & 83.13 & \textbf{82.09} & \textbf{79.10} & \textbf{78.32} & \textbf{77.05} & \textbf{78.35} & \textbf{79.27} & \textbf{78.05} & 75.11 & 76.71 & \textbf{82.00} & \textbf{80.18} & \textbf{79.27} & \textbf{78.95}\\
Alternative (PR, CC, KC, AD) & \textbf{83.25} & 81.86 & 78.48 & 77.06 & 76.74 & 78.12 & 79.17 & 77.75 & \textbf{75.29} & \textbf{76.80} & 81.65 & 79.51 & 79.09 & 78.52 \\
\bottomrule
\end{tabular}
}
\end{table*}

\begin{figure}[!tbp]
	\centering
	\includegraphics[width=\linewidth,trim=0 10 0 110, clip]{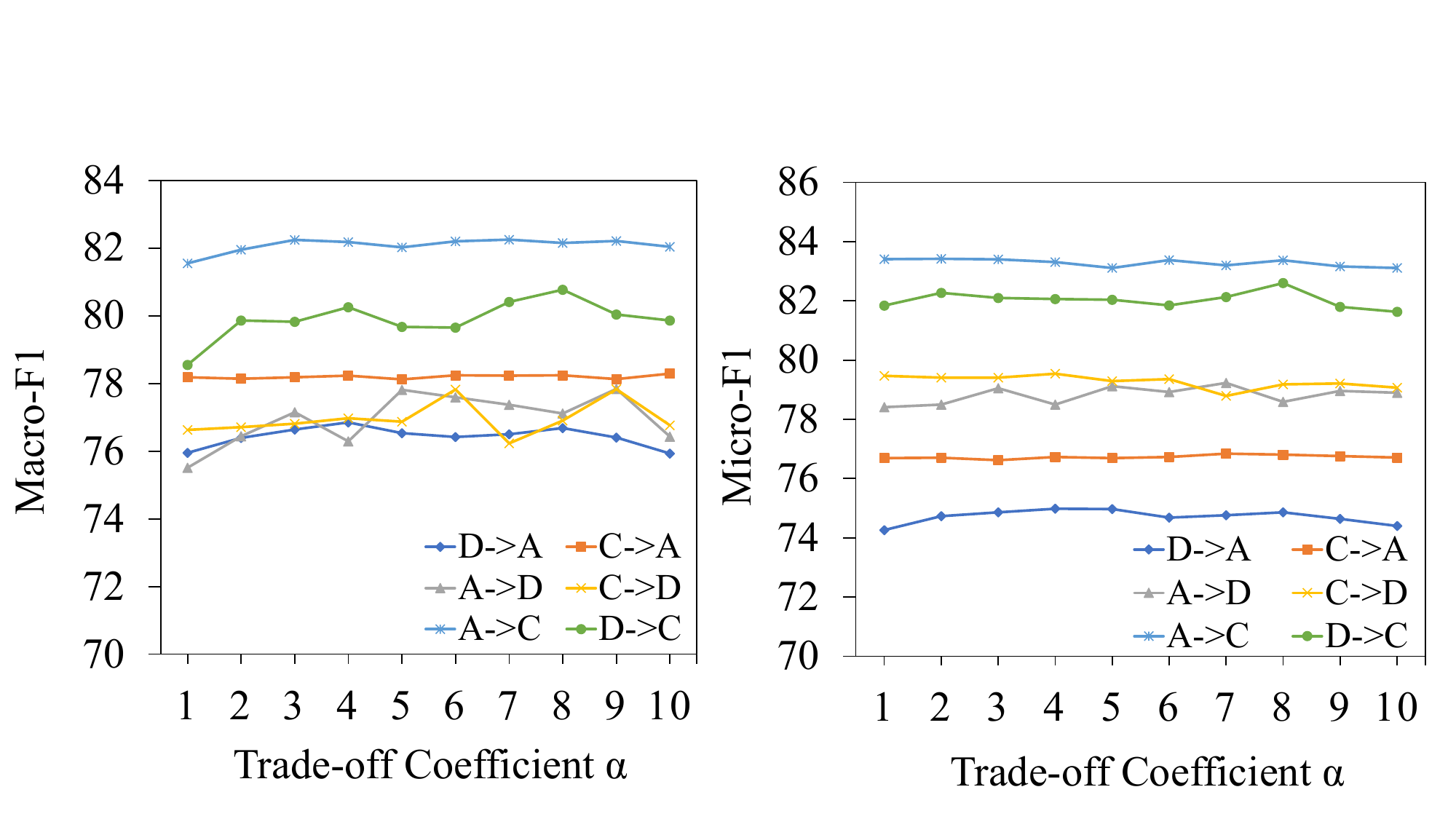}
	\caption{Sensitivity of Micro-F1 and Macro-F1 scores under different weights of $\alpha$ across six cross-domain adaptation scenarios.}
    \label{fig:sensitivity_analysis_weight_alpha}
\end{figure}

\vspace{1ex}
\noindent
\textbf{Impact of trade-off coefficient $\alpha$ of MMD loss}. 
We explore the parametric sensitivity of $\alpha$ across six distinct transfer scenarios, where $\alpha$ governs the regularization strength of cross-domain alignment.
For this analysis, we vary $\alpha$ within the list of $[1, 2, \cdots, 10]$ to observe its comprehensive impact, based on our SNIP with the A2GNN backbone (i.e., SNIP$_{\text{A2GNN}}$).
As illustrated in Fig.~\ref{fig:sensitivity_analysis_weight_alpha},
when $\alpha$ is set to 0, the cross-domain alignment mechanism is totally removed, the model relies solely on the source graph for supervision, and the objective function simplifies to the vanilla classification loss, i.e., $\mathcal{L}_{\mathrm{total}} = \mathcal{L}_{\mathrm{cls}}$. 
Under this zero-alignment configuration, a notable performance degradation is observed across all scenarios. This sharp decline clearly highlights that minimizing cross-domain distribution discrepancy is necessary for mitigating topological shifts and ensuring robust generalization.

\begin{figure}[!tbp]
	\centering
	\includegraphics[width=\linewidth, trim=0 40 0 80, clip]{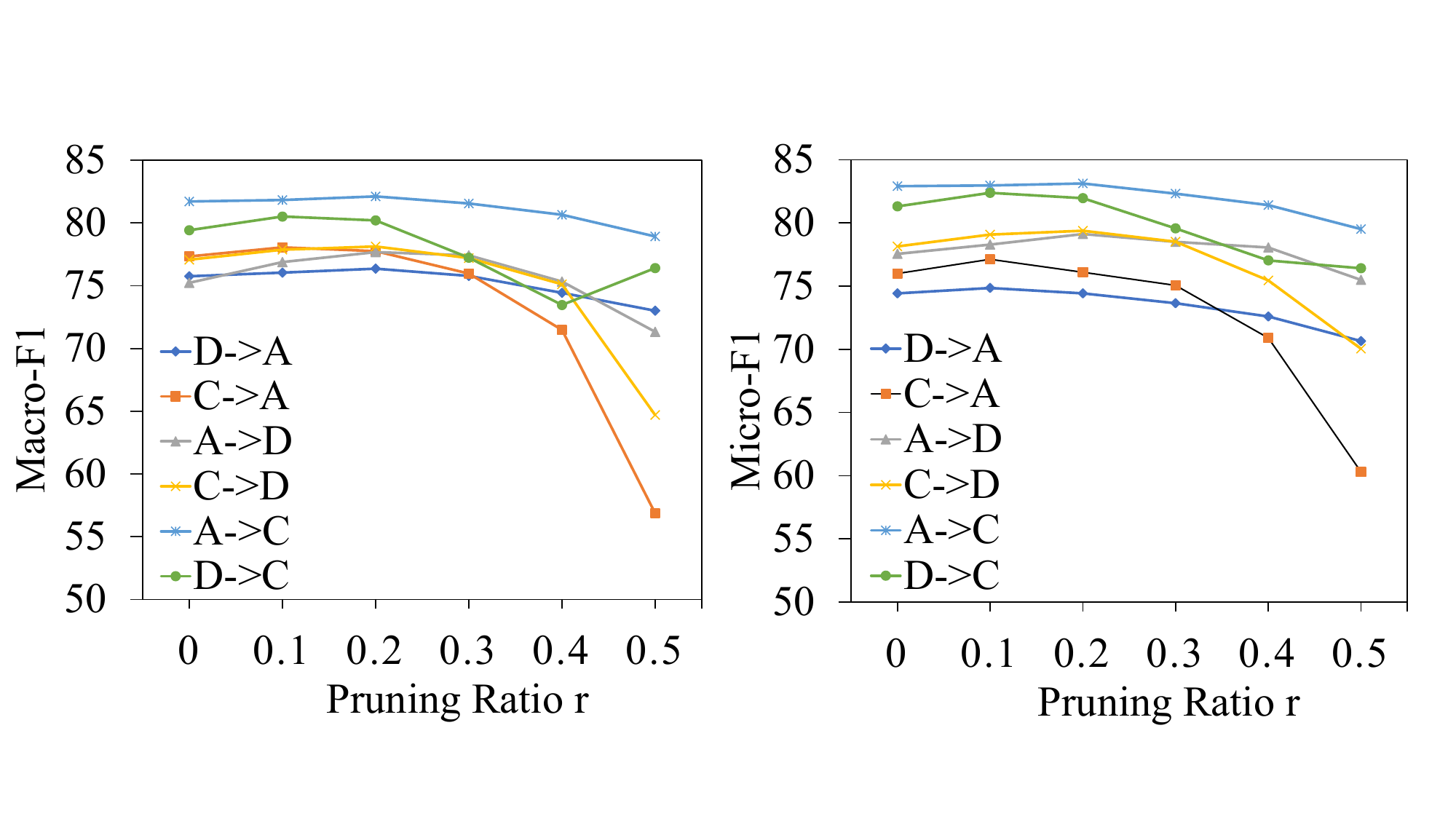}
	\caption{Comparisons of Micro-F1 and Macro-F1 scores with different pruning rates on six cross-domain adaptation scenarios.}
     \label{fig:sensitivity_analysis_pruning_ratio}
\end{figure}

\vspace{1ex}
\noindent\textbf{Impact of Pruning Ratio $r$}. 
We analyze the sensitivity of the model’s performance with respect to the pruning ratio $r$, which controls the proportion of nodes to be removed during the structural pruning process, thereby directly impacting the graph topological structure and the dynamics of information propagation. 
Specifically, $r$ is varied within the list $[0,0.1,0.2,0.3,0.4, 0.5,0.6]$ with a step size of 0.1, and the corresponding performance based on our SNIP with the A2GNN backbone (i.e., SNIP$_{\text{A2GNN}}$) is measured via Micro-F1 and Macro-F1 scores.
As illustrated in Fig.~\ref{fig:sensitivity_analysis_pruning_ratio}, 
when $r < 0.1$, the impact of pruning remains marginal, likely due to the preservation of redundant structural noise. Within the interval $[0.1, 0.3]$, both Micro-F1 and Macro-F1 scores reach a peak performance. 
However, when $r > 0.3$, further increasing the pruning ratio results in a significant performance degradation.  
This degradation indicates that excessive pruning can disrupt the graph's intrinsic connectivity and semantic structure. 
Consequently, these analyses indicate that a pruning rate within the range $[0.1, 0.3]$ strikes an optimal trade-off between noise elimination and structural preservation.

\subsection{Visualization Analysis}
\begin{figure}[!tbp]
	\centering
	\includegraphics[width=\linewidth, trim=0 30 0 40, clip]{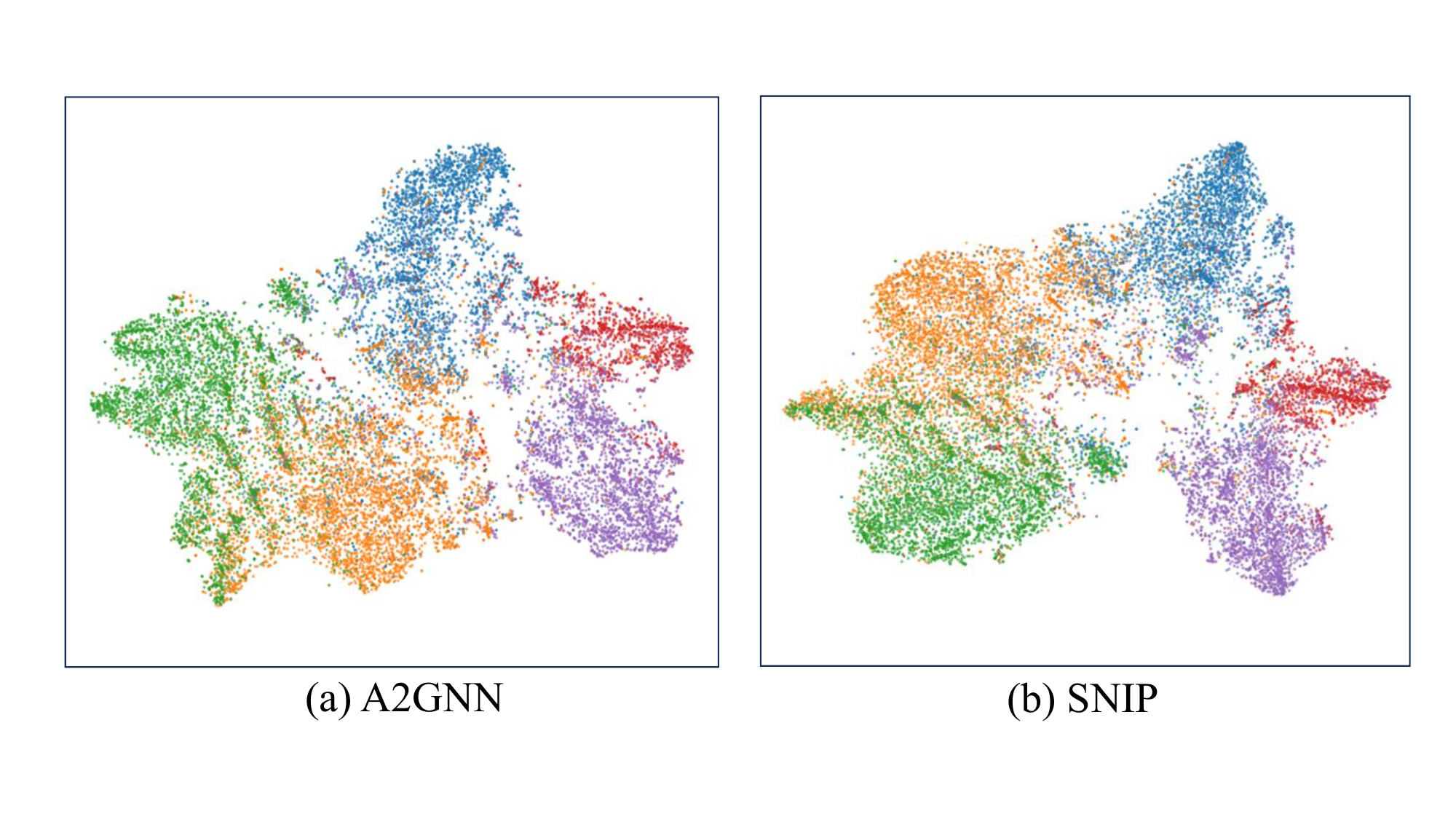}
	\caption{T-SNE visualizations of the learned node representations under the C$\rightarrow$A cross-domain adaptation scenario, comparing our proposed SNIP$_{\text{A2GNN}}$ (right) and the baseline model A2GNN (left).}
    \label{fig:T-SNE}
\end{figure}

To qualitatively examine the distribution alignment and feature discriminability between the source and target domains,
we employ t-SNE \cite{Hinton2002Stochastic} to project the high-dimensional node embeddings into a two-dimensional visual space.
We conduct this evaluation on the challenging C$\rightarrow$A transfer scenario (Citationv1 as the source domain and ACMv9 as the target domain), where each color corresponds to a specific semantic category.
As illustrated in Fig.~\ref{fig:T-SNE}, 
the representations learned by SNIP$_{\text{A2GNN}}$ exhibit significantly clearer decision boundaries and superior cluster separation compared to the baseline A2GNN. 
Notably, our SNIP framework achieves remarkable intra-class cohesiveness and sharper inter-class margins, effectively 
minimizing inter-cluster overlap.
This visualization strongly suggests that SNIP successfully enhances class discriminability within the shared latent embedding space. By filtering out structurally incompatible nodes that typically corrupts domain alignment, SNIP generates superior feature representations and yields robust generalization capabilities.

\section{Conclusion}

In summary, 
this study proposes Source Node Influence Pruning (SNIP), a novel data-refinement framework for Unsupervised Graph Domain Adaptation (UGDA).
SNIP is designed to identify and remove source nodes that are structurally incompatible with the target domain, thereby alleviating cross-domain structural distribution shifts and reducing their adverse effects on knowledge transfer.
Specifically, SNIP first calculates the influence scores of source nodes by integrating multiple centrality-based structural measures.
It then ranks all source nodes in descending order according to their influence scores and removes a predefined proportion of nodes with the lowest scores. By selectively preserving source nodes that are more structurally compatible with the target domain, SNIP constructs a refined source graph that provides a cleaner and more reliable foundation for subsequent feature-level cross-domain alignment, ultimately improving adaptation performance.
Notably, SNIP operates entirely at the data level and is independent of any specific model architecture. It can therefore be integrated into existing UGDA backbones in a seamless and plug-and-play manner, offering strong flexibility and scalability.
Substantial experiments with different backbone networks across multiple transfer scenarios on five real-world datasets confirm that SNIP significantly improves graph domain adaptation performance, effectively mitigating the negative transfer caused by structural anomalies and provide a practical and general solution for cross-domain node classification.

\section*{Acknowledgments}
% This paper is funded by the National Natural Science Foundation of China under grant No. 61976217 and 62306320, 
% the Natural Science Foundation of Jiangsu Province under grant No. BK20231063, and the Open Project Program of the State Key Laboratory of CAD\&CG of Zhejiang University under grant No. A2424.
This work was supported by ``the Fundamental Research Funds for the Central Universities'' under Grant No. 2025QN1154. 
We sincerely appreciate the reviewers’ time, effort, and constructive feedback on this manuscript.

\balance 
\bibliographystyle{IEEEtran}
\bibliography{refer}

\end{document}